\pdfoutput=1
\documentclass[letterpaper]{article} 
\usepackage{aaai24}  
\usepackage{times}  
\usepackage{helvet}  
\usepackage{courier}  
\usepackage[hyphens]{url}  
\usepackage{graphicx} 
\usepackage{natbib}  
\usepackage{caption} 
\usepackage{algorithm}
\usepackage{algorithmic}

\usepackage{newfloat}
\usepackage{listings}
\DeclareCaptionStyle{ruled}{labelfont=normalfont,labelsep=colon,strut=off} 
\floatstyle{ruled}
\newfloat{listing}{tb}{lst}{}
\floatname{listing}{Listing}
\title{Neural Reasoning About Agents' Goals, Preferences, and Actions}
\author{
    Matteo Bortoletto, Lei Shi, Andreas Bulling
}
\affiliations{
    University of Stuttgart, Germany

    \{matteo.bortoletto, lei.shi, andreas.bulling\}@vis.uni-stuttgart.de
}

\newcommand{\model}{IRENE}
\usepackage{mathtools}
\usepackage{siunitx}
\usepackage{subcaption}
\usepackage{multirow}
\usepackage{lscape}
\usepackage{enumitem}
\usepackage{booktabs}

\begin{document}

\maketitle

\begin{abstract}
    We propose the \textit{Intuitive Reasoning Network} (\model{}) -- a novel neural model for intuitive psychological reasoning about agents' goals, preferences, and actions that can generalise previous experiences to new situations. 
    \model{} combines a graph neural network for learning agent and world state representations with a transformer to encode the task context.
    When evaluated on the challenging Baby Intuitions Benchmark, \model{} achieves new state-of-the-art performance on three out of its five tasks -- with up to \SI{48.9}{\percent} improvement.  
    In contrast to existing methods, \model{} is able to bind preferences to specific agents, to better distinguish between rational and irrational agents, and to better understand the role of blocking obstacles. 
    We also investigate, for the first time, the influence of the training tasks on test performance.
    Our analyses demonstrate the effectiveness of \model{} in combining prior knowledge gained during training for unseen evaluation tasks.
\end{abstract}

\section{Introduction}

Common-sense reasoning refers to a broad class of abilities that enable agents to understand basic facts about events, objects, beliefs, or desires~\citep{mccarthy1989artificial}.
Studies in developmental cognitive science have demonstrated that even young infants have these abilities~\citep{needham1993intuitions, gergely1995taking, aguiar19992} while their absence can be linked to developmental disorders, such as autism~\citep{baron1997mindblindness}.
Given its fundamental importance for human social cognition and behaviour, it is imperative that artificial intelligent (AI) agents possess similar capabilities to effectively be understood by humans and in turn, understand them.
This is crucial for a wide variety of applications, such as robotics~\citep{mota2019commonsense} or human-machine collaboration~\citep{conti2022human}. 
Previous research in machine common-sense reasoning has mainly focused on evaluating language processing~\citep{bhagavatula2020abductive, huang2019cosmos, zellers2019hellaswag, bisk2020piqa, sap2019social, sakaguchi2021winogrande} or visual scene understanding~\citep{yi2019clevrer, smith2019modeling, ates2020craft} in a task-specific manner.
More recently, several new benchmarks have been introduced that allow to assess the more general ability of AI systems to reason about unexpected events or situations~\citep{riochet2020intphys, gandhi2021baby, dasgupta2021benchmark, shu2021agent, piloto2022intuitive, weihs2022inflevel}.
Most of them focus on \textit{intuitive physics}~\citep{yi2019clevrer, smith2019modeling, ates2020craft, riochet2020intphys, dasgupta2021benchmark, piloto2022intuitive, weihs2022inflevel, piloto2018probing} in which computational models have to reason about properties and interactions of physical macroscopic objects. 
In contrast, comparably little advances have been achieved in methods for \textit{intuitive psychology}, i.e. common-sense reasoning about other agents' \textit{mental states} from their observed behaviour~\citep{gandhi2021baby, shu2021agent}.

Similar to studies in developmental cognitive science, evaluation of common-sense reasoning abilities in these benchmarks uses a violation of expectation paradigm~\citep[VoE]{baillargeon1987object}.
In a set of familiarisation trials, an observer model has to form an expectation about a particular agent behaviour.
In a subsequent test trial, given the context extracted from the familiarisation trials, the observer has to judge how expected the behaviour of the agent is.
Expectedness is defined as the observer's prediction error  with respect to a ground truth (e.g. agent's next action).
The core idea behind the VoE paradigm is to pair test trials that may not differ much in traditional error metrics but differ in terms of human reasoning.
As such, VoE allows probing AI capabilities by comparing scenarios that humans can differentiate based on those capabilities.

Inspired by behavioural experiments with infants, Gandhi et al.\ have recently introduced the Baby Intuitions Benchmark~\cite[BIB]{gandhi2021baby} -- a set of tasks that require an observer model to reason about agents' goals, preferences, and actions by observing their behaviour in a grid-world environment.
The BIB poses two key challenges.
First, despite being performed in the same grid-world environment, training and evaluation tasks differ in their number and setting. 
Moreover, test trials in the training set present only expected outcomes, requiring observer models to generalise to unseen situations by combining different pieces of knowledge gained during training. 
Second, the benchmark not only challenges models' ability to predict future actions but also to use them to quantify the expectedness of novel situations. 
As such, the BIB can be seen as posing a meta-learning problem~\citep{rabinowitz2018machine} in which an observer model ``learns to learn'' about other agents' behaviour.
While initial models are based on the machine theory of mind network~\citep{rabinowitz2018machine}, the most recent model~\citep[VT]{hein2022comparing} is based on a video transformer~\citep{neimark2021video}.
VT is rather successful in modelling agents' preferences, but it fails in binding them to specific agents. 
Moreover, VT struggles with understanding the role of blocking barriers and that rational agents, in contrast to irrational ones, move efficiently towards their goal.

To address these limitations we introduce the \textit{\textbf{I}ntuitive \textbf{Re}asoning \textbf{Ne}twork} (\model{}) -- a novel neural network for core intuitive psychology.
\model{} uses a graph neural network (GNN) to obtain rich state embeddings by processing graphs extracted from video frames as well as a transformer to encode the familiarisation trials in a context vector.
We show that \model{} achieves new state-of-the-art performance on three out of five reasoning tasks defined on the BIB.
Our model performs particularly well on tasks that existing models~\citep{gandhi2021baby, hein2022comparing} struggle with: binding preferences to specific agents, differentiating between rational and irrational agents, and understanding how to deal with obstacles that block the agents' goal object. 
We also show that for selected tasks, its predictions are in line with infants’ responses collected on a subset of the BIB~\citep{stojnic2022commonsense}.
Finally, for the first time, we investigate the choice of training tasks on generalisation performance.
Our analyses demonstrate our model's ability to combine knowledge gained during training to solve unseen evaluation tasks.
Moreover, we show that training on one type of task does not necessarily improve performance on similar evaluation tasks but that training on unrelated tasks can lead to improvements. 
\footnote{Our project web-page is accessible at \url{https://perceptualui.org/publications/bortoletto24_aaai/}.}

In summary, our contributions are two-fold:
\begin{itemize}
    \item We propose \model{} -- a novel model for intuitive psychology that combines a GNN and a transformer to learn rich state and context representations. \model{} achieves new state-of-the-art performance on three out of five BIB reasoning tasks.
    In particular, it is capable of binding preferences to specific agents and of modelling blocking obstacles and irrational agents better than existing models.
    \item We are first to provide a detailed analysis of the influence of the chosen training tasks on performance. 
    \model{} can achieve new state-of-the-art reasoning performance only when trained on all training tasks, showing its ability to combine knowledge gained during training to solve unseen evaluation tasks.
\end{itemize}

\section{Related Work}

\subsection{Common-Sense Reasoning Benchmarks}

Benchmarks for common-sense reasoning are becoming increasingly popular as they enable the first steps towards intelligent, collaborative agents that reason like humans.
Most research has focused on evaluating language processing~\citep{bhagavatula2020abductive, huang2019cosmos, zellers2019hellaswag, bisk2020piqa, sap2019social, sakaguchi2021winogrande} or visual scene understanding~\citep{yi2019clevrer, smith2019modeling, ates2020craft} through task-specific assessments, e.g.\ predictive accuracy.
However, a number of benchmarks that aim to evaluate the general ability of AI systems to reason about unexpected events or situations have recently emerged~\citep{riochet2020intphys, shu2021agent, gandhi2021baby, dasgupta2021benchmark, piloto2022intuitive, weihs2022inflevel}.
Most of these benchmarks focus on intuitive physics, targeting physical concepts such as continuity, solidity, object persistence and gravity~\citep{smith2019modeling, riochet2020intphys, dasgupta2021benchmark, piloto2022intuitive, weihs2022inflevel}.
Conversely, benchmarks that test models' ability to reason about other agents have received less attention. 
A notable exception is the recent BIB~\citep{gandhi2021baby} that is based on the findings that infants expect other agents to 
have goals, preferences and engage in instrumental actions. 
The benchmark includes a set of tasks to evaluate whether an AI observer can exhibit the same capabilities. 
Like the BIB, AGENT tests whether models can predict that agents have object-based goals and act efficiently~\citep{shu2021agent}. 
In contrast to the BIB, however, AGENT does not evaluate whether models can reason about multiple agents, inaccessible goals, instrumental actions, or distinguish between rational and irrational agents.
Moreover, whilst AGENT's training and evaluation sets present only minor differences and training is done using different leave-out splits, the BIB provides a single canonical split to maximise the evaluation of models' generalisability. 
We evaluate our model on the BIB given its accessibility and relevance as a benchmark for assessing intuitive psychological reasoning.

\begin{figure}[t]
    \centering
    \includegraphics[width=\linewidth]{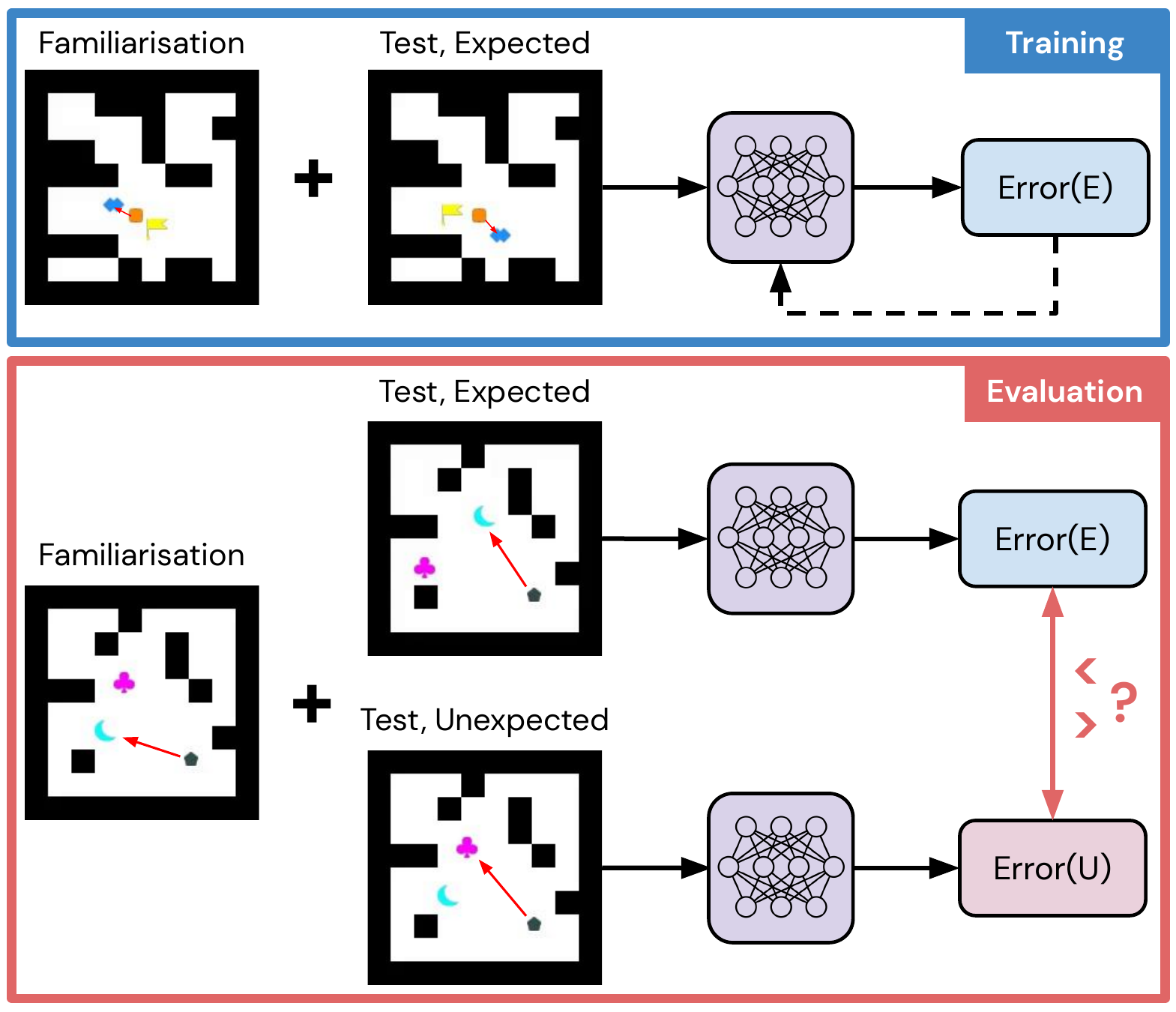}
    \caption{Training and evaluation on the Baby Intuitions Benchmark~\citep{gandhi2021baby}. During training, a model conducts eight familiarisation trials and a test trial, which is always expected.
    The model's weights are updated using backpropagation on the error, computed with respect to the expected test trial ground truth. Evaluation employs a violation of expectation paradigm: based on the familiarisation trials, the model makes predictions on expected and unexpected test trials and both errors are compared. The model is successful if the error on the expected trials is smaller than that on the unexpected trials.}
    \label{fig:bib_vis}
\end{figure}

\subsection{Models for Intuitive Psychology}

Models for reasoning about agents' behaviour and mental states can be grouped into models based on Bayesian theory~\citep{baker2011bayesian, baker2017rational} or deep learning~\citep{rabinowitz2018machine}. 
Shu et al.\ have introduced Bayesian Inverse Planning and Core Knowledge (BIPaCK)~\citep{shu2021agent} that combines Bayesian inverse planning~\citep{baker2017rational} and physics simulation~\citep{battaglia2013simulation}, evaluating it on AGENT. 
More recently, Zhi et al.\ have evaluated a Bayesian Theory of Mind model with hierarchical priors over agents' preference and efficiency (HBToM) on the BIB~\citep{zhi2022solving}.
However, this method makes strong assumptions and uses a tailored definition of expectedness.
Gandhi et al.\ define expectedness in terms of the mean square error between the model prediction and the ground-truth~\citep{gandhi2021baby}. 
Instead, HBToM computes a \emph{plausibility} score by training a set of logistic regression classifiers on a synthetic dataset similar to the BIB evaluation set.
Therefore, in this work we do not compare our results with those of HBToM. 

On the BIB, Gandhi et al.~\citep{gandhi2021baby} have proposed a model based on the Theory of Mind neural network (ToMnet) introduced by~\citep{rabinowitz2018machine}. 
More recently, Hein et al.\ have proposed a method (VT) based on a video transformer~\citep{neimark2021video} that encodes frames using a CNN and performs cross- and self-attention over frames~\citep{hein2022comparing}.
In this work, we introduce a novel method that uses a GNN~\citep{gori2005new} to encode graphs built from frames and a transformer~\citep{vaswani2017attention} to generate context embeddings. 
On AGENT, Shu et al.\ have also used a GNN to encode states, where graphs connect the agent node to all the other nodes~\citep{shu2021agent}. 
Differently, our GNN performs message passing using on heterogeneous graphs with edges representing different spatial relations.

\section{The Baby Intuitions Benchmark}
\label{sec:bib}

The BIB consists of 2D videos of an agent moving and interacting with different objects in a grid-world environment (see Figure~\ref{fig:bib_vis}). 
Both the agent and objects are represented as geometric shapes of different colours.
The benchmark proposes five common-sense reasoning tasks (\textit{Preference}, \textit{Multi-Agent}, \textit{Inaccessible Goal}, \textit{Efficient Action}, and \textit{Instrumental Action}) derived from research on infant intuitive psychology, that require an observer model to reason about agents' goals, preferences, and actions by observing their behaviour in a grid-world environment. The \textit{Efficient Action} and \textit{Instrumental Action} present three sub-tasks each: \textit{Efficiency Path Control}, \textit{Efficiency Time Control}, \textit{Efficiency Irrational Agent}) and \textit{Instrumental No Barrier}, \textit{Instrumental Blocking Barrier}, \textit{Instrumental Inconsequential Barrier}, respectively.
Task descriptions are provided in the Appendix.

An episode in the BIB includes nine trials with trajectories $\{\tau_i\}_{i=1,\dots,9}$, where each trajectory consists of a series of state-action pairs $\tau_i = \{(s_{ij}, a_{ij})\}_{j=1,\dots,T}$, with $s_{ij}$ being video frames and $T$ the trial length. 
$\{\tau_i\}_{i=1,\dots,8}$ are familiarisation trials and $\tau_9$ is the test trial. 
Familiarisation trials serve to give a context to the model while test trials are used to make predictions. 
A test trial can be consistent with the familiarisation examples (expected outcome) or inconsistent (unexpected outcome). 
According to the VoE paradigm, if the observer is more surprised by the unexpected outcome, this means that what they believed or predicted would happen is not in line with what actually occurred.
Expectedness is defined as the observer model's prediction error: a model is successful if the prediction error on the unexpected outcome is higher than the error on the expected outcome. 
In practice, the prediction error is quantified by the mean squared error (MSE) with respect to a ground truth (e.g. next frame or action).

The authors provide a single canonical split for training and evaluation, as they differ in terms of tasks and sample distributions.
The evaluation set presents the tasks mentioned above.
Familiarisation and test trials follow different distributions: expected and unexpected trials are perceptually and conceptually different from familiarisation trials, respectively. 
Consider, for example, the \textit{Preference} task (see Figure~\ref{fig:bib_vis}).
In expected trials, the preferred object is located differently than in familiarisation trials but the agent still moves towards it.
In the unexpected trials, the agent moves towards the non-preferred object whose location is the same as familiarisation.  

The training set presents four tasks: \textit{Single-Object}, \textit{No-Navigation Preference}, \textit{Single-Object Multiple-Agent} and \textit{Agent-Blocked Instrumental Action}.
The training tasks are more trivial and less informative than the evaluation tasks, compelling models to combine and generalise the knowledge acquired from the different training tasks to solve the evaluation tasks. 
For example, for the \textit{Instrumental Blocking Barrier} task, a model has to put together knowledge of navigation (\textit{Single-Object}) and instrumental actions (\textit{Agent-Blocked Instrumental Action}). 
However, at the same time, the model has to generalise its knowledge: for example, in the training trials (\textit{Agent-Blocked Instrumental Action}), the agent is confined in the barrier, whereas in the evaluation trials (\textit{Instrumental Blocking Barrier}) it is the object which is confined. 
For a detailed description of the dataset we refer the interested reader to the original paper~\citep{gandhi2021baby}. 
We provide a brief summary in the Appendix.

\begin{figure*}[t]
    \centering
    \includegraphics[width=\textwidth]{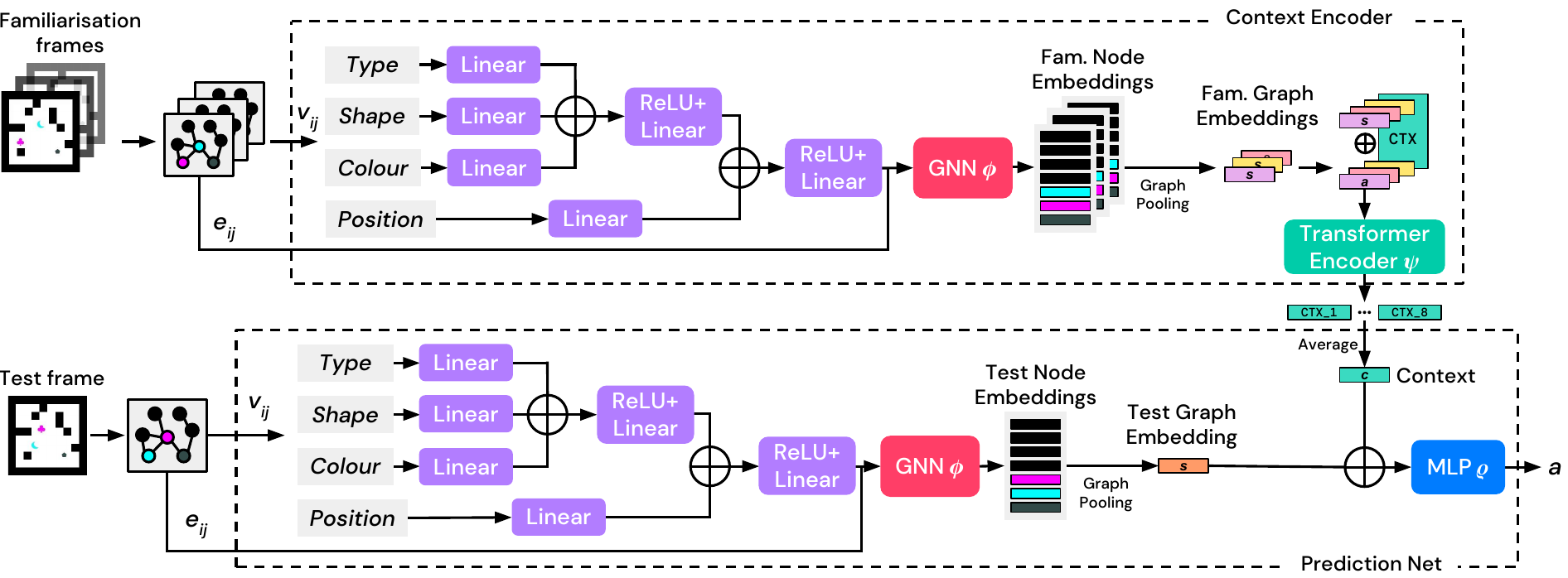}
    \caption{Architecture of \model{}. Inputs are graphs representing entities in a video frame. In the \emph{context encoder}, the state encoder includes a feature fusion module that combines the four node features into a single one, which is input to a GNN. The encoded states are concatenated to the corresponding actions and to a special learnable \texttt{CTX} token.
    A transformer context encoder produces a context embedding as the mean of the \texttt{CXT} embedding vectors of each familiarisation trial. In the \emph{prediction net}, the encoded test state is concatenated to the context and input into a MLP policy that outputs a prediction for the agent's next action. 
    }
    \label{fig:net}
\end{figure*}

\section{Method}
\paragraph{Graph Generation.}

In the BIB, each video is paired with a json file containing information about the grid-world layout, which we use to build graphs.
Videos are sampled at \SI{3}{FPS} and a graph $G_{ij} = (V_{ij}, E_{ij})$ is built from each sampled frame $j=1,\dots,T$ in each trial $i=1,\dots,9$, where $V_{ij}$ is a set of nodes and $E_{ij}$ a set of edges. 
Each entity in a frame is assigned to a graph node $v_{ij}\in V_{ij}$, which has four features: \texttt{type} (e.g.\ ``agent''), \texttt{position} ($x$ and $y$ coordinates), \texttt{colour} (GBR channels) and \texttt{shape} (e.g.\ ``pentagon''). 
We represent categorical variables (\texttt{type} and \texttt{shape}) as one-hot vectors while \texttt{position} and \texttt{colour} are normalised between $[-1,1]$ and $[0,1]$, respectively.
Edges $e_{ij} \in E_{ij}$ represent spatial relationships and are defined following~\citep{jiang2021grid}. 
Specifically, we consider \emph{local directional relations}, which identify the relative position of two adjacent entities, \emph{remote directional relations}, which do not require adjacency, and two \emph{non-directional relations}, adjacent and aligned (see the Appendix for formal definitions). 

\paragraph{Model Architecture.} The architecture of our \textit{Intuitive Reasoning Network} (\model{}) is shown in Figure~\ref{fig:net}.
A \emph{context encoder} parses the agent's past trajectories (i.e. state-action pairs) into a context vector, and a \emph{prediction net} predicts the future behaviour of the agent based on the context and the current state. 
In the state encoder, a feature fusion module combines the node features (\texttt{type}, \texttt{position}, \texttt{colour}, \texttt{shape}). 
For each node $v_{ij}\in V_{ij}$, each feature $v_{ij}^k$ is first embedded using a linear layer $f_k$. 
Then, \texttt{type}, \texttt{colour}, and \texttt{shape} are concatenated and passed through a first fusion layer consisting of a ReLU activation~\citep{fukushima1975cognitron, nair2010rectified} and a linear layer $f_{F_1}$. 
The output is concatenated with the \texttt{position} embedding and passed through the second fusion layer, analogous to the first:
\begin{equation}
    \label{eq:1}
    v_{ij}^{t, s, c} = f_{F_1}(\mathrm{ReLU}
    (\bigoplus_{
        \substack{
        k=type,\\ shape, colour}
    } 
    f_k(v_{ij}^k))),
\end{equation}
\begin{equation}
    \label{eq:2}
    v_{ij}' = f_{F_2} ( \mathrm{ReLU}(f_{position}(v_{ij}^{position}) \oplus v_{ij}^{t, s, c}) ),
\end{equation}
where $\oplus$ indicates concatenation. 
The resulting vector constitutes the input feature for the GNN that performs message passing to update the node embeddings and produces a state embedding $h_{ij}$ by applying a final graph average pooling operation:
\begin{equation}
    \label{eq:3}
    h_{ij} = \mathrm{AvgPooling}(\phi(v'_{ij}, e_{ij})).
\end{equation}
As different edges represent different relations, $\phi$ is a Relational GNN~\citep{schlichtkrull2018modeling} which uses different weights for each edge type. 
In particular, we use GraphSAGE layers with LSTM aggregation~\citep{hamilton2017inductive}.

In the context encoder, the state encoder outputs an embedding vector $h_{ij}$ for each frame graph $G_{ij}$, obtained by applying average pooling to the nodes.  
Then, the encoded states $\{h_{ij}\}_{j=1,\dots,T}$ are concatenated to the corresponding actions $\{a_{ij}\}_{j=1,\dots,T}$ and projected to the transformer input dimension by a linear layer $f_{proj}$. 
A learnable $\texttt{CTX}$ token is concatenated to each embedding vector. 
Thereafter, positional embedding is added, followed by layer normalisation~\citep{ba2016layer}.
The result is input into a transformer encoder $\psi$:
\begin{equation}
    \texttt{CTX}'_i = \psi\left(f_{proj}(\{h_{ij}\}_{j=1}^T \oplus \{a_{ij}\}_{j=1}^T) \oplus \texttt{CTX}_{i}\right)
\end{equation}
and the output $\texttt{CTX}'_i$
are taken as trial representations. 
The context encoder outputs a single context embedding obtained by computing the mean of the eight familiarisation trial representations, 
\begin{equation}
   c=\frac{1}{8}\sum_{i=1}^8 \texttt{CTX}'_i. 
\end{equation} 
In the prediction net, a test frame graph $G_{9,j} = (V_{9j}, E_{9j})$ is encoded by the same state encoder (Eq. 1, 2, 3).
The resulting state embedding $h_{9j}$ is concatenated to the context embedding $c$ and input into an MLP policy $\rho$ that outputs the next action prediction:
\begin{equation}
    a_{pred} = \rho(c \oplus h_{9j}),
\end{equation}
which in our case is represented by the agent's next position in the grid-world.

\section{Experiments}

\begin{table}[t]
    \centering
    \resizebox{\linewidth}{!}{
    \setlength\tabcolsep{3pt}
    \begin{tabular}{lcccccc}
        \toprule
        \textbf{Evaluation Task} & \textbf{BC-MLP} & \textbf{BC-RNN} & \textbf{Video-RNN} & \textbf{VT} & \textbf{\model{}}  \\
        \midrule
        Preference & $26.3$ & $\mathit{48.3}$ & $47.6$ & $\mathbf{80.8}$ & $48.5$ \\
        Multi-Agent & $48.7$ & $48.2$ & $\underline{50.3}$ & $\mathit{49.2}$ & $\mathbf{74.9}$ \\
        Inaccessible Goal & $76.9$ & $\mathit{81.6}$ & $74.0$ & $\underline{85.5}$ & $\mathbf{85.8}$ \\
        \midrule 
        Eff.\ Path Control & $94.0$ & $92.8$ & $\mathbf{99.2}$ & $\mathit{97.5}$ & $\underline{98.1}$ \\
        Eff.\ Time Control & $99.1$ & $99.1$ & $\underline{99.9}$ & $\mathit{99.7}$ & $\mathbf{100.0}$ \\
        Eff.\ Irrational Agent & $\underline{73.8}$ & $\mathit{56.5}$ & $50.1$ & $34.1$ & $\mathbf{85.7}$ \\
        \midrule
        Eff.\ Action Average & $\underline{88.8}$ & $82.5$ & $\mathit{83.1}$ & $77.1$ & $\mathbf{94.7}$ \\
        \midrule
        Inst.\ No Barrier & $\underline{98.8}$ & $\underline{98.8}$ & $\mathbf{99.7}$ & $\mathit{97.9}$ & $78.4$ \\
        Inst.\ Incons.\ Barrier & $55.2$ & $\underline{78.2}$ & $\textit{77.0}$ & $\mathbf{91.9}$ & $52.4$ \\
        Inst.\ Blocking Barrier & $47.1$ & $56.8$ & $\mathit{62.9}$ & $64.2$ & $\mathbf{83.5}$ \\
        \midrule
        Inst.\ Action Average & $67.0$ & $\mathit{77.9}$ & $\underline{79.9}$ & $\mathbf{84.7}$ & $71.5$ \\
        \bottomrule
    \end{tabular}
    }
    \caption{VoE accuracy of existing models and \model{} on the BIB evaluation set. 
    Best score is in \textbf{bold}, second best score is \underline{underlined}, third best score is \textit{italic}. 
    }
    \label{tab:results}
\end{table}

\subsection{Technical Details}

\model{}'s feature fusion module encodes the node features using linear layers of hidden dimension \num{96}. 
The state encoder consists of two GraphSAGE layers for each relation, with hidden dimension \num{96} and ELU activation~\citep{clevert2015fast}.
The transformer encoder 
consists of a stack of six layers with four attention heads, feedforward dimension \num{512} and GELU activations~\citep{hendrycks2016gaussian}.
The prediction net uses the same GNN and feature fusion module used in the context encoder. The MLP policy has hidden dimensions \num{256}, \num{128} and \num{256} and output dimension two, corresponding to the $(x,y)$ coordinates of the agent in the next frame. 
Additional training details are reported in the Appendix.

\subsection{Results}\label{sec:results}

We compare \model{} with the three models originally proposed by Gandhi et al.~\citep{gandhi2021baby} -- BC-MLP, BC-RNN, and Video-RNN -- as well as the more recent VT model by Hein et al.~\citep{hein2022comparing}. 
The VoE accuracy scores on all evaluation tasks, averaged over three different runs, are shown in Table~\ref{tab:results}.
Given that results did not vary much across different runs, we only report the errors for all evaluations in the Appendix.
In line with previous work~\citep{gandhi2021baby, hein2022comparing}, we calculated expectedness as the maximum prediction error. 
We additionally evaluated our model using the mean prediction error and report these results in the Appendix.
As can be seen from Table~\ref{tab:results}, our model achieves state-of-the-art results on three out of five tasks (\textit{Multi-Agent}, \textit{Inaccessible Goal}, \textit{Efficient Action}). 
Moreover, when also considering the \textit{Efficient Action} and \textit{Instrumental Action} sub-tasks, \model{} achieves state-of-the-art results on five out of nine tasks. 
We conducted t-tests to compare IRENE's performance with the baselines.
All results were significant ($\alpha = 0.05$, $p < 0.01$) with only two exceptions: \textit{Preference} between IRENE and BC-RNN; \textit{Time Control} between IRENE and Video-RNN.
The larger improvements are in the \textit{Multi-Agent}, \textit{Instrumental Blocking Barrier} and \textit{Efficiency Irrational Agent} tasks. 
In the \textit{Multi-Agent} task, \model{} dramatically outperforms the other models, improving over the previous best score (Video-RNN) by \SI{48.9}{\percent}.
In the \textit{Instrumental Blocking Barrier} sub-task, our model improves by \SI{30}{\percent} on the previous best score (VT).
In the \textit{Efficiency Irrational Agent} task, \model{} outperforms BC-MLP by \SI{16}{\percent}.
This results in an improvement in the \textit{Efficient Action} task of \SI{6.6}{\percent} with respect to the previous best model (BC-MLP). 
Similar to the baseline methods, our model performs well on the \textit{Path Control} and \textit{Time Control} sub-tasks. 
Remarkably, the score on the \textit{Time Control} sub-task is perfect. 
Scores on the \textit{No Barrier} and \textit{Inconsequential Barrier} tasks are lower than those of the other methods. 
Overall, our model struggles the most in the \textit{Preference} (\num{48.5}) and \textit{Instrumental Action} tasks (average \num{71.5}), especially in the \textit{Inconsequential Barrier} (\num{52.4}). 

\begin{table}[t]
    \centering
    \resizebox{\columnwidth}{!}{
    \begin{tabular}{lcccccc}
        \toprule
        \textbf{BIB Task} & \textbf{LSTM} & \textbf{GCN} & \textbf{Local} & \textbf{Remote} & \textbf{\model{}} \\
        \midrule
        Preference & $48.2$ & $49.7$ & $49.8$ & $50.7$ & $48.5$ \\
        Multi-Agent & $49.7$ & $50.3$ & $98.2$ & $50.0$ & $79.4$ \\
        Inaccessible Goal & $84.8$ & $58.1$ & $41.1$ & $80.6$ & $85.8$ \\
        \midrule 
        Eff.\ Path Control & $97.3$ & $94.7$ & $31.7$ & $98.2$ & $98.1$ \\
        Eff.\ Time Control & $99.9$ & $98.5$ & $37.6$ & $99.8$ & $100.0$ \\
        Eff.\ Irrational Agent & $52.4$ & $89.3$ & $99.7$ & $83.6$ & $85.7$ \\
        \midrule
        Eff. Action Average & $83.2$ & $94.2$ & $56.3$ & $93.9$ & $94.7$ \\
        \midrule
        Inst.\ No Barrier & $78.5$ & $64.6$ & $51.6$ & $78.7$ & $78.4$ \\
        Inst.\ Incons.\ Barrier & $53.3$ & $52.1$ & $52.4$ & $52.7$ & $52.4$ \\
        Inst.\ Blocking Barrier & $83.2$ & $48.0$ & $48.9$ & $83.8$ & $83.5$ \\
        \midrule
        Inst. Action Average & $71.7$ & $54.8$ & $51.0$ & $71.7$ & $71.5$ \\
        \bottomrule
    \end{tabular}
    }
    \caption{
    VoE accuracy for ablated versions of \model{}. 
    ``LSTM'' makes use of an LSTM context encoder instead of the transformer; ``GCN'' substitutes GraphSAGE with GCN layers; ``Local'' takes as input relational graphs with only local directional relations; and ``Remote'' takes as input relational graphs with only remote directional relations.
    }
    \label{tab:ablations}
\end{table}

\subsection{Ablation Studies}

To investigate how different components of our method contribute to these performance improvements, we performed a series of ablation studies summarised in Table~\ref{tab:ablations}. 

\paragraph{Graph Relations.} 
We trained \model{} on graphs whose edges represent only local or remote directional relations. 
Using only local directional relations, the performance on the \textit{Multi-Agent} and \textit{Efficiency Irrational Agent} tasks improved to an almost perfect score. 
However, performance on other tasks became worse, especially in the \textit{Time} and \textit{Path Control} sub-tasks that other models solved almost perfectly. 
This decrease in performance was to be expected as using local relations alone leaves many nodes isolated, including the agent's node. 
As a consequence, the message passing is lacking important information, such as the presence of obstacles that are not adjacent to the agent or to the objects. 
Using only remote relations, performance is comparable to the original model except for \textit{Multi-Agent}, which is at chance level. 
In combination, this suggests that global relations contribute more to the final scores than local ones.

\paragraph{State Encoder.} 
We replaced the GraphSAGE layers with GCN layers in the state encoder. 
The resulting model achieved considerably worse scores in the \textit{Multi-Agent}, \textit{Inaccessible Goal}, \textit{Instrumental No Barrier} and \textit{Blocking Barrier} (sub-)tasks. 
This suggests that GraphSAGE is more effective for modelling obstacles and, as such, is key for our model to better understand the role of blocking barriers and to bind preferences to specific agents.

\paragraph{Context Encoder.} 
We also replaced the transformer encoder $\psi$ with an LSTM. 
Performance remained mostly unchanged except for the \textit{Multi-Agent} and \textit{Efficiency Irrational Agent} tasks where performance dropped to chance level.
This is in line with~\citep{gandhi2021baby} who showed that models with an LSTM failed to adapt their predictions according to whether an agent was rational or irrational during familiarisation.
In combination with the results obtained when substituting GraphSAGE with GCN, this shows that good performance on the \textit{Multi-Agent} task can be obtained only by including both GraphSAGE and a transformer in the model architecture.

\subsection{Analysis of the Training Tasks}
\label{sec:analysis_background_training}

\begin{figure}[t]
    \centering
    \includegraphics[width=\linewidth]{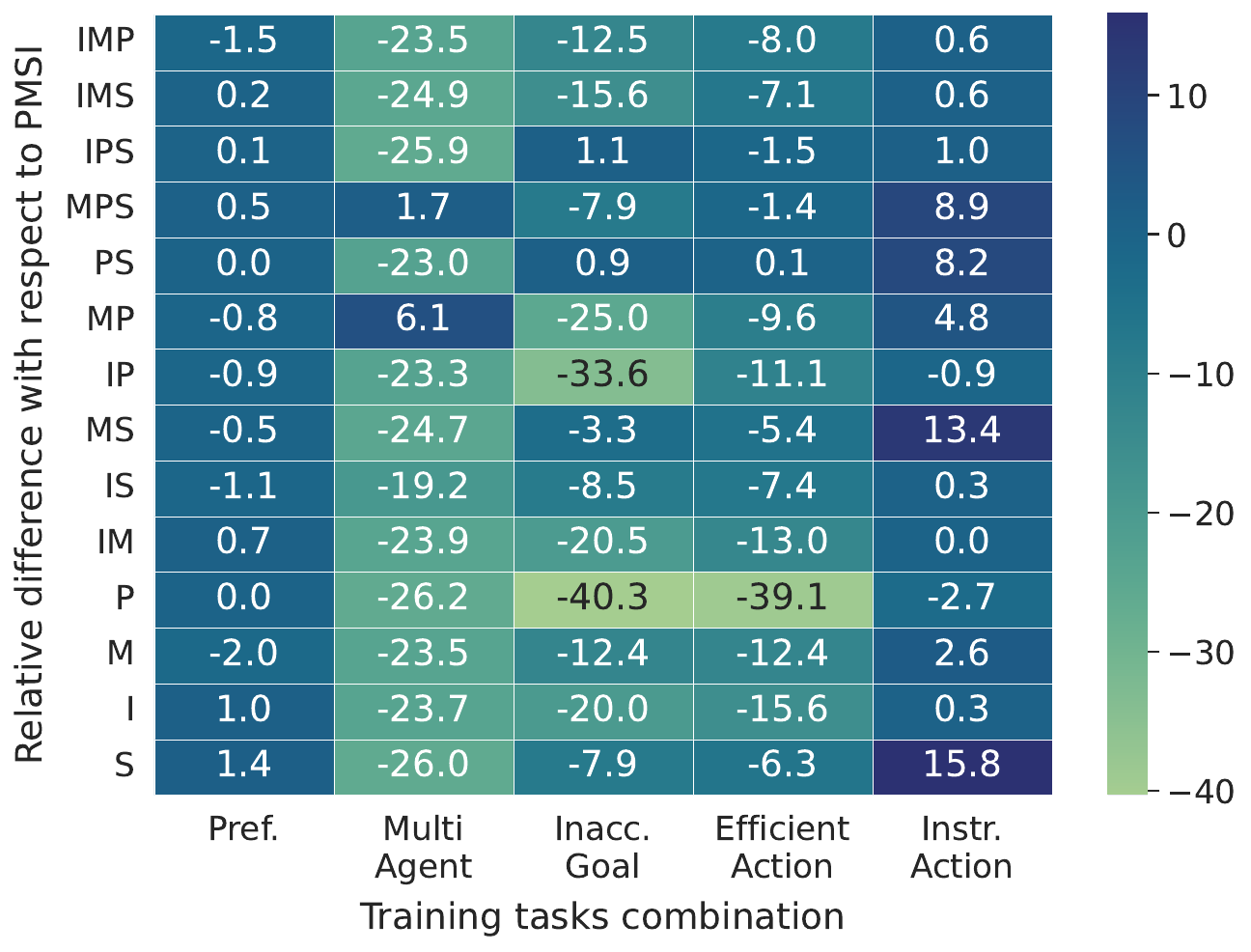}
    \caption{Relative difference of VoE scores obtained on the evaluation set by training on all possible combinations of training tasks with respect to training on all tasks.}
    \label{fig:training-analysis}
\end{figure}

\begin{figure}[t]
    \centering
    \includegraphics[width=\linewidth]{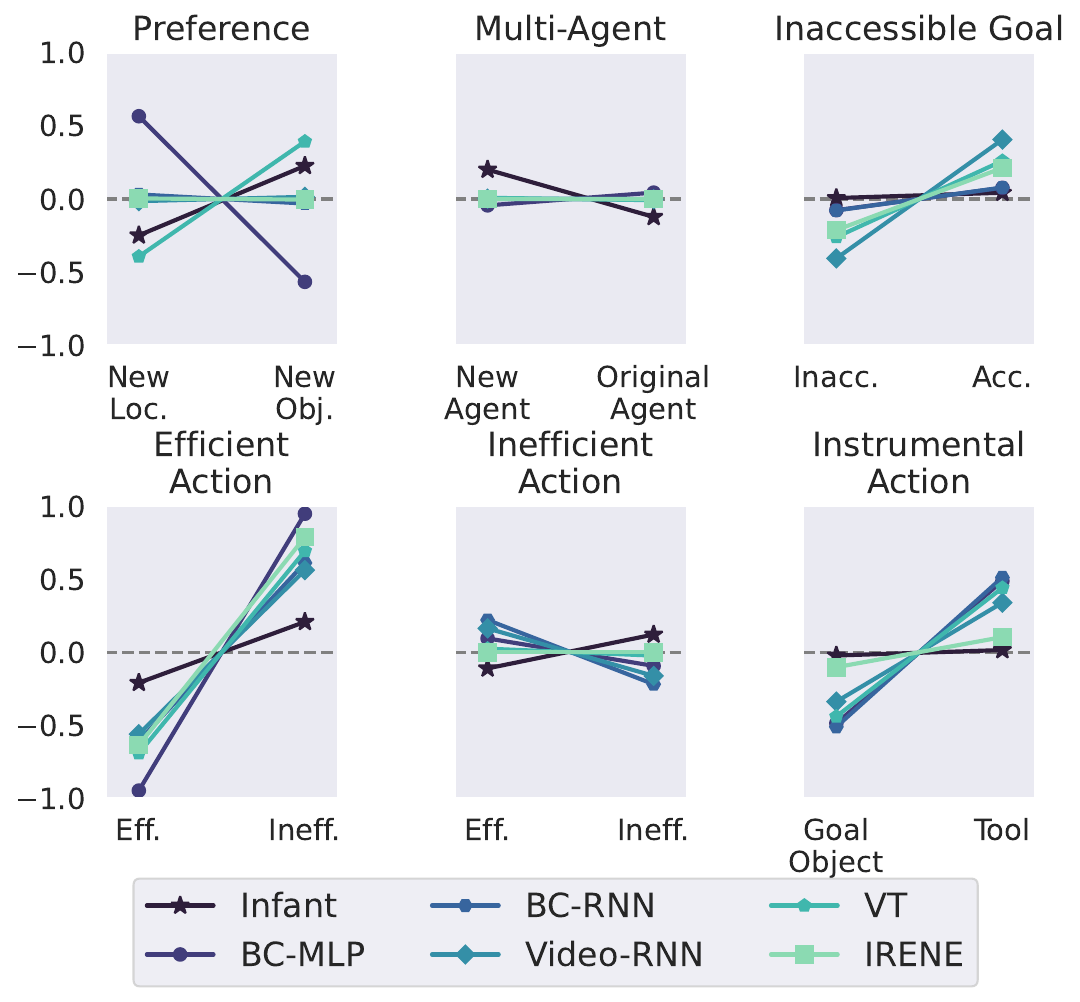}
    \caption{Z-scored means of infants' looking times and models' scores for expected and unexpected outcomes in the BIB evaluation episodes. Positive values indicate expectedness, negative values indicate unexpectedness.}
    \label{fig:z-scored}
\end{figure}

Prior work on the BIB has focused on improving performance~\citep{gandhi2021baby, hein2022comparing} while the question of if and how the choice of training task(s) impacts performance remains under-explored.
This is surprising given that, similar to human learning~\citep{clarke2010teachers}, it is reasonable to assume that computational models also benefit more from some training tasks than others.
To fill this gap, we performed the first investigations into the influence of training tasks on evaluation performance.
To this end, we trained \model{} on the four individual tasks as well as on all of their possible combinations. 
To indicate a combination of training tasks, we used the following notation: P = \textit{No-Navigation Preference}, M = \textit{Single-Object Multi-Agent}, I = \textit{Agent-Blocked Instrumental Action}, S = \textit{Single-Object}. 
For example, IMP indicates joint training on \textit{Agent-Blocked Instrumental Action}, \textit{Single-Object Multi-Agent}, and \textit{No-Navigation Preference} tasks.

Relative differences in performance scores compared to training on all tasks are shown in Figure~\ref{fig:training-analysis} 
(absolute scores are reported in the Appendix). 
Training only on a subset of tasks generally leads to a decrease in performance -- in some cases drastic.
This demonstrates the importance but also the effectiveness of \model{} in extracting and combining knowledge gained from different training tasks.
One notable exception is when training on MPS: in this case the total average performance is comparable to full training ($75.1$ vs.\ $75.4$). 
In particular, the score on the \textit{Inaccessible Goal} task is worse ($77.7$) while the one on the \textit{Instrumental Action} task is better ($80.4$), despite the model not having been trained on I. 
Such improvement is due to better scores on sub-tasks in which blocking barriers are absent or irrelevant: \textit{No Barrier} ($84.8$) and \textit{Inconsequential Barrier} ($92.7$).

Training on S improved scores both on the \textit{No Barrier} and \textit{Inconsequential Barrier} sub-tasks, resulting in a higher score in the \textit{Instrumental Action} task. 
When training on M, performance on the \textit{Multi-Agent} task is considerably worse and training on P does not lead to improvements in the \textit{Preference} task.
This is in line with the general observation that training on one type of task does not always improve performance for similar types of tasks in the evaluation set.

When trained on a subset of the tasks, performance scores show a similar pattern. 
When \model{} is trained on P and M but not on I (i.e.\ MPS, MP), performance on \textit{Multi-Agent} is good.
Considering each evaluation task, while performance on the \textit{Preference} task remains unchanged, it degrades notably for \textit{Multi-Agent}, \textit{Inaccessible Goal}, and \textit{Efficient Action} in most cases.
The score on the \textit{Instrumental Action} task increased for selected combinations of training tasks (MPS, PS, MS, S), especially in MS and S. 

\subsection{Comparison to Infants' Intuitions}

Figure~\ref{fig:z-scored} shows a comparison between the z-scored means of infants' looking times as collected by~\citet{stojnic2022commonsense} and the corresponding models' scores for expected and unexpected evaluation trials. 
As can be seen from the figure, our model's expectations generally align with those of the infants, specifically for the \textit{Inaccessible Goal}, \textit{Efficient Action}, \textit{Inefficient Action}, and \textit{Instrumental Action} tasks.
\model{} performs better in the \textit{Inefficient Action} 
task in which the other models are more surprised at the expected outcome than the unexpected one.
One notable outlier is \textit{Multi-Agent} for which infant behaviour differs from all models.
As reported by \citep{stojnic2022commonsense}, this can be explained by infants not always reacting as expected.

\section{Discussion}
Our experiments demonstrate the effectiveness of our model in addressing intuitive psychology tasks.
\model{} outperforms the state of the art for three out of five reasoning tasks on the challenging Baby Intuitions Benchmark -- in some cases with significant improvements (see Table~\ref{tab:results}). 
More specifically, \model{} better learns to bind preferences to specific agents and to model the role of blocking barriers.
This suggests that, in contrast to existing models~\citep{gandhi2021baby, hein2022comparing}, \model{} relies less on heuristics, such as directly moving towards the goal object. 
However, this also leads to lower scores on sub-tasks for which such simple heuristics are sufficient and which, consequently, allow models that employ them to perform better.
One example of this is the \textit{Instrumental Action} tasks, where models that did not learn the role of barriers during training apply a simple heuristic -- i.e.\ directly moving towards the goal object -- that works on the \textit{Instrumental No Barrier} and \textit{Inconsequential Barrier} sub-tasks but not on the more challenging \textit{Blocking Barrier} sub-task. 
In contrast, IRENE has proven to be effective in handling complex tasks that cannot be tackled by heuristics alone.
This highlights the importance of training models that can truly reason, rather than just applying shortcuts or heuristics, and resonates with the well known problem of reward hacking in deep reinforcement learning~\citep{amodei2016concrete}.
IRENE represents an important advancement on this challenging endeavour.
Our model also obtains near-perfect or perfect scores on the \textit{Path Control} and \textit{Time Control} sub-tasks, demonstrating that it can effectively find the shortest path to an object goal. 
The improved score on the \textit{Efficient Action} task suggests that \model{} can also better model rational agents' behaviour.
In particular, in the \textit{Irrational Agent} sub-task, our model can distinguish between rational agents that move efficiently towards their goal and irrational agents that are not as efficient. 
The \textit{Irrational Agent} is also the task in which the \model{} z-scored mean is the closest to the infants' (see Fig.~\ref{fig:z-scored}).

Our ablation studies (see Table~\ref{tab:ablations}) demonstrate the effectiveness of the proposed combination of GraphSAGE for agent and world state encoding with a transformer for task context encoding. 
We speculate GraphSAGE performs well thanks to inductivity: learning an aggregator allows the model to effectively generate embeddings for nodes which the model only sees during the evaluation trials.
In addition, the non-sequential nature of the transformer together with its self-attention mechanism allow it to overcome LSTM limitations.
We also performed an extensive analysis to better understand the importance of the chosen training tasks on evaluation performance.
Our results confirm the requirements set by \citet{gandhi2021baby} who have argued that models have to combine knowledge from different training tasks:
\model{} performs best when trained on all training tasks while training only on a subset of tasks generally leads to a decrease in performance (see Fig.~\ref{fig:training-analysis}). 
Exceptions are mainly due to a lack of knowledge of blocking barriers, which results in the effective heuristic of ignoring them altogether. 

\paragraph{Limitations and Future Work.} 
By working on \model{} we identified several points that we believe are crucial for future work.
First, \model{} does not perform better on all tasks -- just like the previous models we compared it to.
This suggests that further advances are still needed. 
To fully solve the BIB tasks, it may also be necessary to learn complementary basic concepts, such as from intuitive physics.
Second, besides notable exceptions such as the BIB, advances on intuitive psychology are currently slowed down by the lack of accessible and well-maintained benchmarks.
In particular, we would have liked to evaluate our model on AGENT \citep{shu2021agent} -- that covers an interesting environment and complementary set of challenging reasoning tasks -- but AGENT does not provide any benchmark or model code.
Despite our significant efforts, re-implementing the method and reproducing the results based solely on the information provided in the paper turned out to be an extremely challenging task.
Likely because of this also no other papers have evaluated on AGENT to date.
There is an urgent need for the community to design and create new benchmarks to foster the development of ``general neural common-sense reasoners''.
Third, despite the general trend of designing benchmarks to be used only for evaluation~\citep{weihs2022inflevel}, future work should also explore how introducing new training tasks allows models to learn more effectively.
This resonates with our suggestion to train models on data representing different basic concepts.

\section{Conclusion}
In this work we have proposed \model{} -- a novel neural network for reasoning about agents’ goals, preferences, and actions.
IRENE sets new state-of-the-art on three out of five tasks on the Baby Intuitions Benchmark, with improved modelling of agents' preferences, obstacle handling, and distinguishing between rational and irrational agents.
We also demonstrated the effectiveness of IRENE in combining knowledge gained during training for unseen evaluation tasks.
These results are not only promising for advancing human-like reasoning in AI systems but also shed new light on the importance of the choice of training tasks for good generalisation performance.

\section{Acknowledgements}
M. Bortoletto and A. Bulling were funded by the European Research Council (ERC) under the European Union's Horizon 2020 research and innovation programme under grant agreement No 801708.
L. Shi was funded by the Deutsche Forschungsgemeinschaft (DFG, German Research Foundation) under Germany's Excellence Strategy -- EXC 2075 -- 390740016.
The authors would like to especially thank Pavel Denisov, Hsiu-Yu Yang, Ekta Sood, and Manuel Mager for numerous insightful discussions, and thank Ann-Sophia M\"{u}ller and Constantin Ruhdorfer for their technical support.

\bibliography{refs}

\appendix

\section{Appendix}

\subsection{BIB Dataset Details} \label{app:bib}

The BIB evaluation set contains different numbers of videos for different tasks. 
Videos have been recorded at \SI{25}{FPS} with a resolution of $200\times200$ pixels. 
The five common-sense reasoning tasks in the evaluation set are the following:
\begin{itemize}
    \item \textit{Preference} task: understand that an agent has a preferred goal object;
    \item \textit{Multi-Agent} task: bind preferences to specific agents;
    \item \textit{Inaccessible Goal} task: understand the role of physical obstacles and predict that an agent will approach another object if the preferred one is not accessible;
    \item \textit{Efficient Action} task:
    this task consists of three sub-tasks. Learn that a rational agent will move efficiently towards its goal in terms of (a) spatial distance (\textit{Efficiency Path Control}) and (b) time (\textit{Efficiency Time Control}), as well as (c) learn to differentiate between rational and irrational agents (\textit{Efficiency Irrational Agent)};
    \item \textit{Instrumental Action} task: understand that agents remove blocking barriers if it allows them to attain a higher-level goal. The task consists of three sub-tasks, depending on whether the barrier is (a) absent (\textit{Instrumental No Barrier}), (b) blocking the goal object (\textit{Instrumental Blocking Barrier}), or (c) not blocking it (\textit{Instrumental Inconsequential Barrier}).
\end{itemize}
The training set presents four tasks that are more trivial and less informative than the evaluation tasks: 
\begin{itemize}
    \item \textit{Single-Object} task: an agent moves towards an object at different locations in the grid-world;
    \item \textit{No-Navigation Preference} task: an agent moves towards a preferred object among two, with both objects very close to the agent's initial location;
    \item \textit{Single-Object Multiple-Agent} task: similar to the Single-Object Task but at a designed point in time a second agent takes the initial agent's place;
    \item \textit{Agent-Blocked Instrumental Action} task: an agent is trapped within a blocking barrier. To remove the barrier and reach its goal, the agent first has to collect a key and insert it into the barrier lock.
\end{itemize}
This compels models to combine and generalise the knowledge acquired from the different training tasks to solve the evaluation tasks.

Specifically, we used the evaluation set version 1.1, which contains the following number of samples:
\begin{itemize}
    \item Preference: 1,000 videos;
    \item Multi-Agent: 1,000 videos;
    \item Inaccessible Goal: 1,111 videos;
    \item Efficient Action:
    \begin{itemize}
        \item Efficiency Path Control: 1,500 videos;
        \item Efficiency Time Control: 1,000 videos;
        \item Efficiency Irrational Agent: 1,000 videos
    \end{itemize}
    \item Instrumental Action:
    \begin{itemize}
        \item Instrumental No Barrier: 999 videos;
        \item Instrumental Blocking Barrier: 327 videos;
        \item Instrumental Inconsequential Barrier: 330 videos.
    \end{itemize}
\end{itemize}
The training set contains:
\begin{itemize}
    \item Single-Object: 8,000 videos;
    \item No-Navigation Preference: 7,985 videos;
    \item Single-Object Multiple-Agent: 3,200 videos;
    \item Agent-Blocked Instrumental Action: 3,200 videos.
\end{itemize}
Each video is paired with a json file containing information about the grid-world layout, which we use to build frame graphs.  

\begin{table*}[t]
    \centering
    \begin{tabular}{lccccc}
        \toprule
        \textbf{Evaluation Task} & \textbf{BC-MLP} & \textbf{BC-RNN} & \textbf{Video-RNN} & \textbf{VT} & \textbf{\model{}}  \\
        \midrule
        Preference & $26.3$ & $\mathit{48.3}$ & $47.6$ & $\mathbf{80.8}\pm 0.0$ & $\underline{48.5}\pm0.0$ \\
        Multi-Agent & $48.7$ & $48.2$ & $\underline{50.3}$ & $\mathit{49.2}\pm 0.0$ & $\mathbf{74.9}\pm0.0$ \\
        Inaccessible Goal & $76.9$ & $\mathit{81.6}$ & $74.0$ & $\underline{85.5}\pm 0.0$ & $\mathbf{85.8}\pm0.0$ \\
        \midrule 
        Eff.\ Path Control & $94.0$ & $92.8$ & $\mathbf{99.2}$ & $\mathit{97.5}\pm 0.0$ & $\underline{98.1}\pm0.0$ \\
        Eff.\ Time Control & $99.1$ & $99.1$ & $\underline{99.9}$ & $\mathit{99.7}\pm0.0$ & $\mathbf{100.0}\pm0.0$ \\
        Eff.\ Irrational Agent & $\underline{73.8}$ & $\mathit{56.5}$ & $50.1$ & $34.1\pm 0.1$ & $\mathbf{85.7}\pm0.4$ \\
        \midrule
        Eff.\ Action Average & $\underline{88.8}$ & $82.5$ & $\mathit{83.1}$ & $77.1\pm0.0$ & $\mathbf{94.7}\pm0.1$ \\
        \midrule
        Inst.\ No Barrier & $\underline{98.8}$ & $\underline{98.8}$ & $\mathbf{99.7}$ & $\mathit{97.9}\pm 0.0$ & $78.4\pm0.0$ \\
        Inst.\ Incons.\ Barrier & $55.2$ & $\underline{78.2}$ & $\textit{77.0}$ & $\mathbf{91.9}\pm 0.0$ & $52.4\pm0.0$ \\
        Inst.\ Blocking Barrier & $47.1$ & $56.8$ & $\mathit{62.9}$ & $\underline{64.2}\pm 0.1$ & $\mathbf{83.5}\pm0.4$ \\
        \midrule
        Inst.\ Action Average & $67.0$ & $\mathit{77.9}$ & $\underline{79.9}$ & $\mathbf{84.7}\pm0.0$ & $71.5\pm0.1$ \\
        \bottomrule
    \end{tabular}
    \caption{VoE accuracy of existing models and \model{} on the BIB evaluation set. 
    Best score is in \textbf{bold}, second best score is \underline{underlined}, third best score is \textit{italic}. 
    }
    \label{tab:results_errors}
\end{table*}

\subsection{Graph relations} \label{app:graph-rel}

Based on \citep{jiang2021grid}, we built a graph for each video frame with edges representing different spatial relations. Given two entities $a$ and $b$ in the grid-world, local directional relations are defined as 
\begin{align*}
    \mathrm{RightAdj}(a,b) &\leftarrow (x_a = x_b + l) \land (y_a = y_b), \\
    \mathrm{LeftAdj}(a,b) &\leftarrow (x_a = x_b - l) \land (y_a = y_b), \\
    \mathrm{TopAdj}(a,b) &\leftarrow (y_a = y_b + l) \land (x_a = x_b), \\
    \mathrm{BottomtAdj}(a,b) &\leftarrow (y_a = y_b - l) \land (x_a = x_b), \\
    \mathrm{TopRightAdj}(a,b) &\leftarrow (x_a = x_b + l) \land (y_a = y_b + l), \\
    \mathrm{TopLeftAdj}(a,b) &\leftarrow (x_a = x_b - l) \land (y_a = y_b + l), \\
    \mathrm{BottomRightAdj}(a,b) &\leftarrow (x_a = x_b + l) \land (y_a = y_b - l), \\
    \mathrm{BottomLeftAdj}(a,b) &\leftarrow (x_a = x_b - l) \land (y_a = y_b - l). 
\end{align*}
where $l$ is the spacing of the grid.
Remote directional relations are defined as 
\begin{align*}
    \mathrm{Right}(a,b) &\leftarrow x_a > x_b, \\
    \mathrm{Left}(a,b) &\leftarrow x_a < x_b, \\
    \mathrm{Top}(a,b) &\leftarrow y_a > y_b, \\
    \mathrm{Bottom}(a,b) &\leftarrow y_a < y_b.
\end{align*}
Alignment and adjacency relations are defined as
\begin{align*}
    \mathrm{Aligned}(a,b)  &\leftarrow (x_a = x_b) \lor (y_a = y_b), \\
    \mathrm{Adjacent}(a,b) &\leftarrow (\lvert x_a - x_b \rvert \leq l) \land (\lvert y_a - y_b \rvert \leq l).
\end{align*}

\subsection{Technical Details}
\label{app:train-details}
We trained \model{} for 32 epochs using the Adam optimiser~\citep{kingma2014adam} with $\beta_1=0.9$, $\beta_2=0.99$, $\epsilon=10^{-8}$ and learning rate $\eta = 5\cdot10^{-4}$.
\SI{80}{\percent} of the training episodes were used for training while the rest were used for validation. 
We selected hyperparameters empirically, i.e.\ did not perform any exhaustive hyperparameter tuning.
All experiments were conducted on a single NVIDIA Tesla V100 GPU with 32 GB VRAM and a batch size of \num{32}.

We trained our models using PyTorch~\citep{paszke2019pytorch} with \texttt{7}, \texttt{42} and \texttt{123} as random seeds. 
We performed our data analysis using NumPy~\citep{harris2020array}, Pandas~\citep{reback2020pandas, mckinney2010data}, and SciPy~\citep{virtanen2020scipy}. 
Figures were made using Matplotlib~\citep{Hunter2007matplotlib}.

\subsection{VoE Scores}
VoE accuracy scores and error of existing models and IRENE on the BIB evaluation set are reported in Table~\ref{tab:results_errors}.
As discussed in the Discussion section, in contrast to existing methods, IRENE relies less on heuristics, such as directly moving towards the goal object. 
This leads to lower scores on sub-tasks for which simple heuristics are sufficient. 
For example, let us consider the Instrumental Action tasks. 
In such tasks, models that did not learn the role of barriers during training apply a simple heuristic -– i.e. directly moving towards the goal object –- that works on the \textit{Instrumental No Barrier} and \textit{Inconsequential Barrier} sub-tasks but not on the more challenging \textit{Blocking Barrier} sub-task. 
In contrast, IRENE is effective for complex tasks that cannot be tackled by heuristics alone. 
This highlights the importance of training models that can truly reason, rather than just applying shortcuts or heuristics.

\subsection{Expectedness: Max vs.\ Mean.} 

\begin{table*}[t]
    \centering
    \begin{tabular}{lcccccccc}
        \toprule
        \textbf{Evaluation Task} & \textbf{VT} & \textbf{VT} (mean) & \textbf{\model{}} & \textbf{\model{}} (mean) \\
        \midrule
        Preference & $80.8\pm 0.0$ & $82.1\pm0.0$ & $48.5\pm0.0$ & $49.6\pm0.0$ \\
        Multi-Agent & $49.2\pm 0.0$ & $49.1\pm0.0$ & $74.9\pm0.0$ & $63.6\pm0.5$ \\
        Inaccessible Goal & $85.5\pm 0.0$ & $89.8\pm0.0$ & $85.8\pm0.0$ & $88.3\pm0.1$ \\
        \midrule 
        Eff.\ Path Control & $97.5\pm 0.0$ & $97.3\pm0.0$ & $98.1\pm0.0$ & $97.3\pm0.0$ \\
        Eff.\ Time Control & $99.7\pm0.0$ & $99.8\pm0.0$ & $100.0\pm0.0$ & $100.0\pm0.0$ \\
        Eff.\ Irrational Agent & $34.1\pm 0.1$ & $29.5\pm0.1$ & $85.7\pm0.4$ & $81.2\pm0.6$ \\
        \midrule
        Eff.\ Action Average & $77.1\pm0.0$ & $75.5\pm0.0$ & $94.7\pm0.1$ & $92.9\pm0.2$ \\
        \midrule
        Inst.\ No Barrier & $97.9\pm 0.0$ & $98.7\pm0.0$ & $78.4\pm0.0$ & $83.8\pm0.0$ \\
        Inst.\ Incons.\ Barrier & $91.9\pm 0.0$ & $96.9\pm0.0$ & $52.4\pm0.0$ & $52.4\pm0.0$ \\
        Inst.\ Blocking Barrier & $64.2\pm 0.1$ & $82.1\pm0.1$ & $83.5\pm0.4$ & $99.4\pm0.0$ \\
        \midrule
        Inst.\ Action Average & $84.7\pm0.0$ & $92.6\pm0.0$ & $71.5\pm0.1$ & $78.5\pm0.0$ \\
        \bottomrule
    \end{tabular}
    \caption{Performance of VT~\citep{hein2022comparing} and \model{} on the BIB evaluation set. Scores evaluate VoE judgements in which expectedness is expressed as the maximum or mean (when specified) prediction error.
    }
    \label{tab:results_mean}
\end{table*}

Gandhi et al.\ have defined the expectedness of a test trial as the model's maximum prediction error on the trial frames~\citep{gandhi2021baby}.
The authors justified their choice by claiming that alternatives, like using the mean, consistently resulted in lower scores. 
In contrast, Hein et al.\ have obtained higher scores when using the mean~\citep{hein2022comparing}. 
Table~\ref{tab:results_mean} shows that for \model{}, using the mean leads to comparable scores to using the maximum for most of the tasks, with the exception of \textit{Instrumental Blocking Barrier} (improvement), \textit{Multi-Agent} (deterioration) and \textit{Efficiency Irrational Agent Agent} (deterioration).
To be comparable to the baseline models, in the main text we report results using the maximum prediction error. 

\subsection{Ablation Studies -- Absolute scores} \label{app:ablations}
Table~\ref{tab:app-abl} shows absolute scores with errors for our analysis of the training tasks. 

\begin{table*}[t]
    \centering
    \begin{tabular}{lcccccc}
        \toprule
        \textbf{BIB Task} & \textbf{LSTM} & \textbf{GCN} & \textbf{Local} & \textbf{Remote} & \textbf{\model{}} \\
        \midrule
        Preference & $48.2\pm0.0$ & $49.7\pm0.0$ & $50.0\pm0.2$ & $50.7\pm0.0$ & $48.5\pm0.0$ \\
        Multi-Agent & $49.7\pm0.0$ & $50.3\pm0.2$ & $98.0\pm0.2$ & $50.0\pm0.0$ & $74.9\pm0.0$ \\
        Inaccessible Goal & $84.8\pm0.2$ & $58.1\pm0.3$ & $41.7\pm0.5$ & $71.6\pm0.0$ & $85.8\pm0.0$ \\
        \midrule 
        Eff.\ Path Control & $97.3\pm0.0$ & $94.7\pm0.0$ & $31.8\pm0.1$ & $90.8\pm0.1$ & $98.1\pm0.0$ \\
        Eff.\ Time Control & $99.9\pm0.0$ & $98.5\pm0.0$ & $37.2\pm0.3$ & $99.3\pm0.5$ & $100.0\pm0.0$ \\
        Eff.\ Irrational Agent & $52.4\pm0.1$ & $89.3\pm0.1$ & $99.4\pm0.3$ & $79.2\pm0.8$ & $85.7\pm0.4$ \\
        \midrule
        Eff. Action Average & $83.2\pm0.0$ & $94.2\pm0.0$ & $56.1\pm0.2$ & $93.9\pm0.5$ & $94.7\pm0.1$ \\
        \midrule
        Inst.\ No Barrier & $78.5\pm0.0$ & $64.6\pm0.2$ & $50.7\pm0.8$ & $78.4\pm0.3$ & $78.4\pm0.0$ \\
        Inst.\ Incons.\ Barrier & $53.3\pm0.0$ & $52.1\pm0.0$ & $52.8\pm0.3$ & $52.7\pm0.0$ & $52.4\pm0.0$ \\
        Inst.\ Blocking Barrier & $83.2\pm0.4$ & $48.0\pm0.3$ & $45.6\pm0.0$ & $83.1\pm0.6$ & $83.5\pm0.4$ \\
        \midrule
        Inst. Action Average & $71.7\pm0.1$ & $54.8\pm0.2$ & $49.7\pm0.4$ & $71.4\pm0.3$ & $71.5\pm0.1$ \\
        \bottomrule
    \end{tabular}
    \caption{VoE scores for ablated versions of \model{}. 
    ``LSTM'' makes use of an LSTM context encoder instead of the transformer; ``GCN'' substitutes GraphSAGE with GCN layers; ``Local'' takes as input relational graphs with only local directional relations; and ``Remote'' takes as input relational graphs with only remote directional relations.
    }
    \label{tab:app-abl}
\end{table*}

\subsection{Analysis of the Training Tasks -- Absolute scores} \label{app:background-train-analysis}
Table~\ref{tab:training-analysis} shows absolute scores with errors for our analysis of the training tasks.

\subsection{Analysis of the Training Tasks -- Relation Between Training Tasks} 
The relation between different training task combinations and performance change (see Figure 3 in the main text) raises some interesting discussion points, which we report in the following.  

\paragraph{Training on M, MP, and IMP.}
We noticed that training on M only does not boost performance for \textit{Multi-Agent}, but training on MP does. Yet, training on IMP again decreased the performance on M. 

Training on M does not improve performance for \textit{Multi-Agent} because, when training on M, the preferred object is always located close to the agent’s initial location. 
In these training samples the agent hardly moves and when our model is exclusively trained on M, it thus fails to grasp the dynamics of how agents should move. 
Consequently, when tested in a \textit{Multi-Agent} scenario where the agent needs to navigate and reach preferred objects situated at various locations in the gridworld, the performance suffers.

However, training on MP increases performance. This is directly related to the first point. When our model is exclusively trained on M, it fails to grasp the dynamics of how agents should move. 
However, in P the agent moves to various locations of the grid, allowing our model to learn how agents move. 
This results in a boost in performance.

IMP again decreased the performance on \textit{Multi-Agent}.
This happens because I is unrelated to \textit{Multi-Agent}. 
In I, an agent is trapped within a blocking barrier. 
To remove the barrier and reach its goal, the agent first has to collect a key and insert it into the barrier lock. 
In the \textit{Multi-Agent} task the agent is not trapped and there is no key or barrier. 
Therefore, performance on \textit{Multi-Agent} never benefits from training on I. 

\paragraph{Training on MPS} 
We also noticed that training on MPS causes a decrease in \textit{Inaccessible Goal} and an increase in \textit{Instrumental Action}. 
In the \textit{Inaccessible Goal} task, during the familiarisation trials, the agent navigates to its preferred object in various locations. 
In the expected test trials, the preferred object becomes inaccessible, causing the agent to go to the non-preferred object.
In unexpected test trials the preferred object is accessible but the agent goes to the non-preferred object. 
Training on MPS leads to a decrease in performance on the \textit{Inaccessible Goal} task because in MPS trials the preferred object is always accessible, and the agent can always reach it.
The increase in performance on the \textit{Instrumental Action} task is due to a similar reason. 
The \textit{Instrumental Action} task comprises three sub-tasks. 
Two of these sub-tasks, \textit{Instrumental No Barrier} and \textit{Inconsequential Barrier}, can be solved without requiring knowledge about barriers (the test trials in \textit{Instrumental No Barrier} even lack barriers) by just applying simple heuristics. 
Consequently, a model trained on MPS performs well in these two tasks but poorly in the more challenging \textit{Instrumental Blocking Barrier} task. 
This performance pattern results in a higher average score compared to a model that performs better in the challenging \textit{Blocking Barrier} task but not as well in the other two.

\begin{landscape}
\begin{table}[p]
    \setlength{\tabcolsep}{3pt}
    \centering
    \resizebox{1.28\textheight}{!}{
    \begin{tabular}{lccccccccccccccc}\hline
    \toprule
    \multirow{2}[2]{*}{\textbf{Evaluation Task}} & \multicolumn{14}{c}{\textbf{Training Tasks}} \\
                                                 \cmidrule{2-16}
                                                 & IMPS & IMP & IMS & IPS & MPS & PS & MP & IP & MS & IS & IM & P & M & I & S \\
    \midrule
    Preference                            & $48.5\pm0.0$ & $47.0\pm0.0$ & $48.7\pm0.0$ & $48.6\pm0.0$ & $49.0\pm0.0$ & $48.5\pm0.0$ & $47.7\pm0.0$ & $47.6\pm0.0$ & $48.0\pm0.0$ & $47.4\pm0.0$ & $49.2\pm0.0$ & $48.5\pm0.0$ & $46.5\pm0.0$ & $49.5\pm0.0$ & $49.9\pm0.0$ \\ 
    Multi-Agent                           & $74.9\pm0.0$ & $51.4\pm0.0$ & $50.0\pm0.0$ & $49.0\pm0.0$ & $76.4\pm0.7$ & $51.9\pm0.0$ & $80.4\pm1.3$ & $51.6\pm0.0$ & $50.3\pm0.2$ & $55.4\pm0.7$ & $50.9\pm0.1$ & $49.9\pm2.6$ & $51.5\pm0.1$ & $51.3\pm0.2$ & $48.9\pm0.0$ \\  
    Inaccessible Goal                     & $85.8\pm0.0$ & $73.6\pm0.2$ & $70.5\pm0.3$ & $87.0\pm0.1$ & $77.7\pm0.2$ & $86.5\pm0.2$ & $60.7\pm0.1$ & $52.1\pm0.1$ & $82.3\pm0.3$ & $77.4\pm0.3$ & $65.3\pm0.1$ & $45.8\pm0.3$ & $73.2\pm0.1$ & $65.3\pm0.3$ & $77.8\pm0.1$ \\  
    \midrule
    Eff.\ Path Control              & $98.1\pm0.1$ & $90.8\pm0.2$ & $98.6\pm0.0$ & $97.5\pm0.0$ & $94.7\pm0.0$ & $95.2\pm0.0$ & $81.7\pm0.0$ & $81.0\pm0.0$ & $97.9\pm0.1$ & $97.8\pm0.0$ & $94.7\pm0.0$ & $49.2\pm0.2$ & $97.3\pm0.1$ & $89.5\pm0.0$ & $98.4\pm0.1$ \\  
    Eff.\ Time Control              & $100.0\pm0.0$ & $98.3\pm0.0$ & $100.0\pm0.0$ & $99.9\pm0.0$ & $100.0\pm0.0$ & $100.0\pm0.0$ & $95.3\pm0.0$ & $92.1\pm0.0$ & $100.0\pm0.0$ & $100.0\pm0.0$ & $99.4\pm0.0$ & $66.4\pm0.1$ & $99.9\pm0.0$ & $94.4\pm0.0$ & $100.0\pm0.0$ \\  
    Eff.\ Irrational Agent          & $85.7\pm0.4$ & $71.1\pm1.3$ & $63.9\pm0.2$ & $81.3\pm0.8$ & $84.9\pm0.6$ & $88.2\pm0.7$ & $77.3\pm0.7$ & $76.0\pm2.3$ & $70.0\pm0.4$ & $64.0\pm0.1$ & $51.1\pm0.9$ & $50.7\pm1.3$ & $49.1\pm0.8$ & $53.3\pm0.5$ & $66.7\pm0.5$ \\  
    \midrule
    Eff.\ Action Average              & $94.7\pm0.1$ & $86.7\pm0.5$ & $87.5\pm0.0$ & $92.9\pm0.3$ & $93.2\pm0.2$ & $94.5\pm0.2$ & $84.7\pm0.2$ & $83.0\pm0.8$ & $89.3\pm0.2$ & $87.3\pm0.0$ & $81.7\pm0.3$ & $55.4\pm0.5$ & $82.1\pm0.3$ & $79.1\pm0.2$ & $88.4\pm0.2$ \\  
    \midrule
    Inst.\ No Barrier              & $78.4\pm0.0$ & $78.9\pm0.0$ & $79.1\pm0.1$ & $79.4\pm0.0$ & $84.8\pm0.2$ & $84.6\pm0.1$ & $80.7\pm0.1$ & $77.1\pm0.2$ & $87.9\pm0.1$ & $78.6\pm0.0$ & $78.4\pm0.2$ & $69.6\pm0.2$ & $79.4\pm0.4$ & $78.5\pm0.1$ & $89.8\pm0.2$ \\  
    Inst.\ Incons.\ Barrier & $52.4\pm0.0$ & $53.6\pm0.0$ & $53.0\pm0.0$ & $53.3\pm0.0$ & $92.7\pm0.3$ & $91.3\pm0.9$ & $90.6\pm0.3$ & $52.4\pm0.0$ & $94.2\pm0.6$ & $52.7\pm0.0$ & $53.0\pm0.0$ & $73.6\pm0.0$ & $91.7\pm1.0$ & $52.1\pm0.0$ & $95.5\pm0.6$ \\  
    Inst.\ Blocking Barrier        & $83.5\pm0.4$ & $83.5\pm0.0$ & $84.1\pm0.0$ & $84.3\pm0.2$ & $63.6\pm0.0$ & $63.7\pm0.2$ & $58.0\pm0.3$ & $82.6\pm0.3$ & $71.6\pm0.0$ & $83.8\pm0.0$ & $82.7\pm0.2$ & $62.7\pm0.3$ & $49.8\pm0.3$ & $84.4\pm0.3$ & $75.8\pm0.0$ \\  
    \midrule
    Inst.\ Action Average           & $71.5\pm0.1$ & $72.0\pm0.0$ & $72.1\pm0.0$ & $72.3\pm0.1$ & $80.4\pm0.1$ & $79.9\pm0.4$ & $76.4\pm0.2$ & $70.7\pm0.1$ & $84.6\pm0.2$ & $71.7\pm0.0$ & $71.4\pm0.1$ & $68.6\pm0.1$ & $73.6\pm0.6$ & $71.7\pm0.1$ & $87.0\pm0.3$ \\  
    \bottomrule
    \end{tabular}
    }
    \caption{VoE scores for all possible combinations of training tasks. We indicate a combination of training tasks with the following notation: P = \textit{No-Navigation Preference}, M = \textit{Single-Object Multi-Agent}, I = \textit{Agent-Blocked Instrumental Action}, S = \textit{Single-Object}. For example, IMP means jointly training on \textit{No-Navigation Preference}, \textit{Single-Object Multi-Agent} and \textit{Agent-Blocked Instrumental Action}. 
    }
    \label{tab:training-analysis}
\end{table}
\end{landscape}

\end{document}